\documentclass[12pt, letterpaper]{article}

\usepackage[margin=2.5cm, bottom=2.5cm, footskip=1.25cm]{geometry}
\usepackage{mathptmx} 
\usepackage[parfill]{parskip}
\usepackage{setspace}
\usepackage{titlesec}
\titleformat{\section}{\normalfont\fontsize{14}{17}\bfseries\MakeUppercase}{}{0pt}{}
\titlespacing*{\section}{0pt}{\baselineskip}{0pt}
\titleformat{\subsection}{\normalfont\fontsize{12}{15}\bfseries}{}{0pt}{}
\titlespacing*{\subsection}{0pt}{\baselineskip}{0pt}
\titleformat{\subsubsection}[runin]{\normalfont\fontsize{12}{15}\itshape}{}{0pt}{}[.]
\titlespacing*{\subsubsection}{0pt}{\baselineskip}{1em}

\usepackage{caption}
\usepackage{graphicx}
\usepackage{booktabs}
\usepackage{fancyhdr}
\usepackage[style=apa]{biblatex}
\usepackage{multirow}
\usepackage{array}
\usepackage{amsmath}

\begin{document}

\begin{center}
    \vspace*{\baselineskip}
    
    {\fontsize{16}{19}\selectfont \textbf{Built Environment Reasoning from Remote Sensing Imagery Using Large Vision--Language Models} \footnote{Published in the Proceedings of the International Conference on Industrialized Construction 2026.}} \\[1\baselineskip]
    
    {\normalfont Dongdong Wang*, Deepak Balakrishnan, Ravi Srinivasan, and Shenhao Wang*} \\[1\baselineskip]
    
    {\fontsize{12}{15}\itshape 
    Department of Urban and Regional Planning, University of Florida \\ 
    M.E. Rinker, Sr. School of Construction Management, University of Florida} \\[0.5\baselineskip]
    
    {\fontsize{10}{12}\itshape dongdongwang@ufl.edu, shenhaowang@ufl.edu}
\end{center}

\vspace{2\baselineskip}

\section*{Abstract}  

This work investigates the use of large language models (LLMs) for tasks in smart cities. The core idea is to leverage remote sensing imagery to characterize the built environment, including design suggestions, constructability assessment, landuse patterns, and risk identification. We examine remote sensing imagery at multiple spatial scales as inputs for multimodal language modeling and evaluate their effects on built-environment-related reasoning. In addition, we compare state-of-the-art LLMs, including InternVL and Qwen, in terms of accuracy and reliability when generating built environment recommendations. The results demonstrate the potential of integrating remote sensing imagery with large language models to assist smart cities and decision-making.

\textbf{Keywords:} Built Environment; Large Language Models; Remote Sensing 

\section{Introduction}

Large language models (LLMs) have emerged as a powerful paradigm for tackling a wide range of challenges across diverse domains, exhibiting strong capabilities in logical reasoning and inference. Building upon this foundation, vision-language models (VLMs) extend these strengths to multimodal settings, demonstrating promising performance in applications such as medical imaging and remote sensing. However, its capabilities in built environment understanding remain largely underexplored, with limited empirical evidence demonstrating its effectiveness, highlighting a critical gap for further investigation.

Recent progress in remote sensing VLMs (RS-VLMs) \parencite{kuckreja_geochat_2024, pang_vhm_2025, wang_geollava-8k_2025, yao2025falcon} have demonstrated promising reasoning capabilities by combining remote sensing imagery and large VLMs. These existing efforts primarily showcase the power of LLMs . While the RS-VLMs demonstrate strong performance in captioning and semantic understanding, their numeric reasoning capabilities remain underexplored, such as evaluating spatial contexts or calculating urban density metrics. 

Quantitative reasoning can facilitate the analysis of built environment metrics, such as building density, land-use composition, and green coverage, revealing quantifiable geospatial patterns. Traditionally, these metrics are computed through time-consuming pipelines combining GIS, remote sensing imagery, and manual annotations, which require substantial domain expertise and limit scalability. In fact, high-resolution remote sensing imagery offers a scalable and structured visual representation of cities, naturally containing rich built environment and geospatial information. Quantitative reasoning via remote sensing imagery is an ideal testbed for grounded quantitative reasoning because it requires diverse reasoning capabilities including but not limited to object detection, proportional estimation, spatial reasoning, and compositional numerical inference.

In this work, we investigate the potential of remote sensing–enabled large language models for quantitative understanding and estimation of the built environment, with the goal of supporting efficient building energy planning. We first develop a large-scale benchmark to systematically evaluate quantitative reasoning through visual question answering, revealing the limitations of zero-shot inference. We then perform domain-specific fine-tuning with limited training data, achieving significant performance improvements in built environment estimation and demonstrating strong potential for scalable energy planning applications. Finally, we conduct sensitivity analyses across key factors to highlight the strengths and limitations of current approaches, providing insights to guide future research in large language model–driven building energy analysis.

The key contributions of this work are:

\begin{itemize}

\item \textbf{Benchmark development.} We design a scalable benchmark to evaluate the potential of large language models for efficient built environment estimation, with applications in building energy planning and optimization. The benchmark primarily focuses on quantitative reasoning tasks supported by remote sensing imagery.

\item \textbf{Comprehensive LLM evaluation.} We conduct an extensive evaluation of state-of-the-art large language models to assess their capability for efficient inference in quantitative understanding of the built environment. The results reveal notable limitations in direct estimation under zero-shot settings.

\item \textbf{Fine-tuning for built environment analysis.} We further perform targeted fine-tuning with limited training data and demonstrate consistent performance improvements across multiple large language models. These findings highlight the potential of data-efficient adaptation for scalable energy planning applications.

\end{itemize}

\section{Related Work}
\label{sec:related}

\subsection{Built Environment for Energy}

Human activity and spatial footprints shape energy consumption patterns across urban built environments \parencite{fathi2020machine}. Energy demand is closely linked to the intensity and configuration of the built environment \parencite{zhang2026multi}. Urban expansion and increased building density raise energy use and intensify the urban heat island effect, leading to higher cooling demand \parencite{wang2018random}. In contrast, green areas help mitigate heat accumulation, though their effectiveness depends on spatial arrangement and integration with surrounding structures \parencite{im2022impact}. Other factors, such as pavement and impervious surfaces, also influence energy use by affecting heat retention and local climate \parencite{im2022impact}. Understanding these interactions supports more efficient infrastructure planning, improved energy management, and sustainable urban development.

\subsection{Remote Sensing}

Remote sensing provides an efficient approach for scaling the interpretation and modeling of human footprints across large spatial extents. Given the wide coverage of study regions, remote sensing imagery enables consistent and region level characterization of infrastructure and human activity patterns, supporting more informed decision making \parencite{wang2025generative}. The analysis of remote sensing imagery has traditionally relied on a range of digital image processing and computer vision techniques. Recently, advances in large language models have introduced new opportunities to enhance this process, with emerging approaches leveraging language driven reasoning to extract and interpret meaningful footprints from imagery \parencite{li2024vision}. These developments offer promising directions for deriving more reliable and scalable representations of built environment characteristics for research and planning applications.

\subsection{Large Vision-Language Models}

The emergence of large language models (LLMs), such as ChatGPT and LLaMA \parencite{touvron2023llama}, has significantly advanced text generation and reasoning capabilities, enabling more sophisticated multimodal understanding. Building on this progress, recent large vision language models extend these capabilities beyond domain specific tasks toward more general problem solving. Models such as Qwen \parencite{bai2023qwen} and InternVL \parencite{chen_internvl_2024} integrate vision encoders and scalable visual backbones with language models to support unified multimodal reasoning, improved cross modal alignment, and long context understanding. These advances have further driven the development of remote sensing foundation models \parencite{kuckreja_geochat_2024, pang_vhm_2025, yao_falcon_2025}, accelerating progress in remote sensing image analysis.

\section{Benchmark Development.}

\begin{figure*}[h]
    \centering
    \includegraphics[width=1\linewidth]{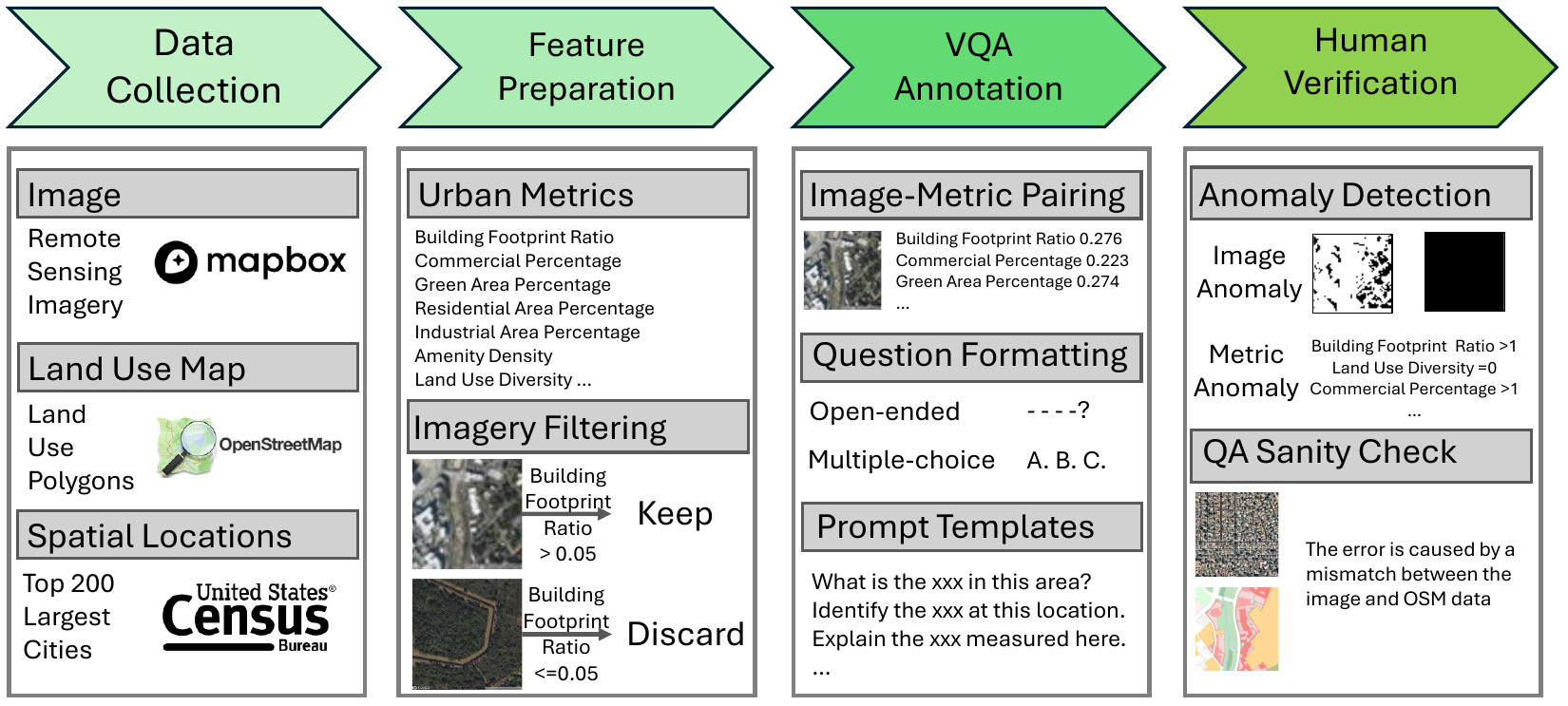}
    \caption{Benchmark preparation pipeline.}
    \label{fig:pipeline}
\end{figure*}

To investigate and demonstrate the potential of large language models for built environment understanding and modeling, we construct a benchmark following a standard pipeline, including data collection, feature engineering, visual question answering generation, and human verification, as illustrated in Figure~\ref{fig:pipeline}.

\subsection{Imagery Data Collection}

We integrate two primary data sources: remote sensing imagery from Mapbox \parencite{rzeszewski2023mapbox} and land use spatial statistics from OpenStreetMap (OSM) \parencite{OpenStreetMap}. Data are organized based on city size using Gazetteer statistics from the U.S. Census Bureau. We select the 335 largest metropolitan areas using city center coordinates from the U.S. Gazetteer Files. These cities provide diverse and dense urban environments, enabling reliable metric derivation while presenting complex scenarios for evaluating vision–language models. For each city, we define a bounding region and extract imagery from Mapbox at zoom level 17 (672 $\times$ 672 pixels), covering approximately 450 m $\times$ 450 m per tile. A 4 $\times$ 4 grid centered at each city yields up to 16 images, capturing surrounding urban contexts. We obtain land use polygons from OSM, including residential, commercial, industrial, and green-related classes (e.g., grass, forest, farmland), as well as water bodies under the natural tag. These spatial annotations are used to derive built environment metrics characterizing urban structure and composition.

\subsection{Built Environment Metrics} 

We carefully select built environment metrics \parencite{santhanam2022quantification, oliveira2020urban} to characterize spatial patterns of urban infrastructure, focusing on factors closely related to energy consumption management and large scale building development. These metrics capture both the structural composition and functional relationships of urban space, including building density, surface coverage, and the balance between natural and built elements (Table~\ref{tab:urban_metrics_summary}). For example, building footprint ratio and impervious space ratio reflect development intensity and surface sealing, which are directly associated with heat retention and energy demand. Green area percentage and urban blue area percentage represent ecological components that contribute to thermal regulation and mitigation of heat accumulation. In addition, ratio based metrics such as blue to built up ratio, green to blue index, and blue to green ratio provide insights into the interactions between built structures and environmental features, enabling a more comprehensive understanding of urban microclimate dynamics. Together, these metrics support systematic analysis of how urban form influences energy consumption and provide a foundation for data driven planning and sustainable infrastructure design.

\begin{table*}[h]
\footnotesize
\centering
\caption{Selected built environment metrics used in this study with definitions.}
\label{tab:urban_metrics_summary}
\setlength{\tabcolsep}{3pt}
\renewcommand{\arraystretch}{1.05}
\begin{tabular}{p{4cm} >{\centering\arraybackslash}p{3cm} p{8.5cm}}
\toprule
\textbf{Metric} & \textbf{Formula} & \textbf{Description} \\
\midrule

Building Footprint Ratio &
$\sum A_i^{\text{footprint}} / A_{\text{total}}$ &
Share of land covered by buildings; development intensity. \\

Green Area Percentage &
$A_{\text{green}} / A_{\text{total}}$ &
Vegetated land share; ecological capacity. \\

Urban Blue Area Percentage &
$A_{\text{blue}} / A_{\text{total}}$ &
Surface water proportion. \\

Impervious Space Ratio &
$A_{\text{impervious}} / A_{\text{total}}$ &
Fraction of impermeable surfaces. \\

Blue/Built Ratio &
$A_{\text{blue}} / A_{\text{built}}$ &
Water relative to built-up land. \\

Green–Blue Index &
$A_{\text{green}} / A_{\text{blue}}$ &
Vegetation relative to water. \\

Blue–Green Ratio &
$A_{\text{blue}} / A_{\text{green}}$ &
Water relative to vegetation. \\

\bottomrule
\end{tabular}
\end{table*}

\subsection{Visual Question Answering} 

The imagery and corresponding metrics are linked through image–question pairs. For each image, multiple questions are generated, each targeting a specific metric for estimation. All questions are designed in a multiple-choice format, where one option is correct and accompanied by three plausible alternatives. The candidate answers are carefully constructed within a reasonable range rather than including extreme outliers, ensuring that the questions remain valid and challenging, focusing on quantitative reasoning rather than simple anomaly detection. Overall, the proposed benchmark dataset contains 14,523 images and 101,661 visual question–answer pairs, with 9,523 images and 66,661 pairs used for training, and 5,000 images and 35,000 pairs reserved for testing. Each image is associated with 7 metrics on average, and the average question length is 8.11 words, providing a concise yet diverse benchmark for evaluating quantitative reasoning capabilities.

\section{Experiments}

\subsection{Training Configuration}
We fine-tune the models using the AdamW optimizer with a learning rate of $5\times10^{-5}$ and a weight decay of 0.01. Training is performed with a batch size of 32 on four NVIDIA B200 GPUs, using gradient accumulation with 8 steps. Models are trained for 2 epochs with a warmup ratio of 5\% of the total training steps, followed by cosine or linear learning rate decay. For parameter-efficient adaptation, we employ LoRA with rank $r=32$ and scaling factor $\alpha=16$, applied to attention layers (i.e., query and value projections). All experiments are implemented in PyTorch using the HuggingFace Transformers library, with mixed-precision (FP16) training.

\subsection{Models}
We selected a diverse set of state-of-the-art VLMs to benchmark their performance on urban imagery reasoning tasks. The general-purpose models include LLaVA-1.6-7B \parencite{liu_improved_2024}, InternVL-3-8B \parencite{chen2024internvl}, QwenVL-3-8B \parencite{bai2025qwen3}, and MGM-7B \parencite{li2025mini} while domain-specific models include GeoChat-7B \parencite{kuckreja_geochat_2024}, VHM-7B \parencite{pang_vhm_2025}, and Falcon-7B \parencite{yao2025falcon}, covering a range of architectures, training paradigms, and multi-modal capabilities. This selection spans models optimized for instruction-following, spatial reasoning, geospatial imagery interpretation, and general multi-step visual reasoning.



\subsection{Performance Results}

\begin{table*}
\centering
\small
\setlength{\tabcolsep}{3pt}  
\caption{
Quantitative reasoning accuracy ($\%$) of VLMs on multiple-choice questions with zero-shot inference.
}
\begin{tabular}{l|cccc|ccc}
\hline
\multirow{2}{*}{Built Environment Metrics} & \multicolumn{4}{c|}{\textbf{General-Purpose Models}} & \multicolumn{3}{c}{\textbf{Remote-Sensing Models}} \\
\cline{2-8}
& LLaVA-1.6 & InternVL-3 & QwenVL-3 & MGM & GeoChat & VHM & Falcon  \\
\hline
Building Footprint Ratio  & 27.71 & 36.21 & 35.21 & 33.41 & 22.14 & 23.09 & 25.55  \\
Green Area Percentage  & 35.06 & 45.72 & 44.18 & 43.96 & 41.29 & 42.97 & 42.01  \\
Urban Blue Area Percentage  & 24.97 & 34.88 & 33.94 & 32.36 & 21.85 & 23.66 & 26.02  \\
Impervious Space Ratio & 25.94 & 36.02 & 35.77 & 34.19 & 22.88 & 23.14 & 25.09  \\
Blue to Built-Up Area Ratio  & 25.02 & 35.61 & 36.41 & 32.79 & 21.56 & 22.84 & 26.57  \\
Green-Blue Infrastructure Index  & 25.23 & 37.12 & 35.08 & 33.58 & 22.26 & 22.95 & 24.91  \\
Blue-Green Ratio & 24.66 & 35.28 & 34.47 & 32.63 & 23.41 & 22.89 & 26.85  \\
\hline
\end{tabular}
\label{tab:sota_accuracy}
\end{table*}

\begin{table*}
\centering
\small
\setlength{\tabcolsep}{3pt} 
\caption{
Quantitative reasoning accuracy ($\%$) of VLMs on multiple-choice questions after domain-specific fine-tuning. 
}
\begin{tabular}{l|cccc|ccc}
\hline
\multirow{2}{*}{Built Environment Metrics} & \multicolumn{4}{c|}{\textbf{General-Purpose Models}} & \multicolumn{3}{c}{\textbf{Remote-Sensing Models}} \\
\cline{2-8}
& LLaVA-1.6 & InternVL-3 & QwenVL-3 & MGM & GeoChat & VHM & Falcon  \\
\hline
Building Footprint Ratio & 51.38 & 62.57 & 58.94 & 56.12 & 44.63 & 47.72 & 49.21  \\
Green Area Percentage & 61.44 & 72.15 & 68.73 & 67.28 & 63.91 & 68.22 & 65.74  \\
Urban Blue Area Percentage & 50.21 & 60.33 & 58.62 & 55.79 & 45.12 & 46.28 & 51.67  \\
Impervious Space Ratio & 48.27 & 60.74 & 61.12 & 57.86 & 49.31 & 48.79 & 50.56  \\
Blue to Built-Up Area Ratio & 49.36 & 59.08 & 64.92 & 56.34 & 45.03 & 48.11 & 52.18  \\
Green-Blue Infrastructure Index & 50.71 & 65.63 & 61.84 & 58.06 & 47.89 & 47.21 & 50.37  \\
Blue-Green Ratio & 49.02 & 59.73 & 58.95 & 54.11 & 49.86 & 47.35 & 53.29  \\
\hline
\end{tabular}
\label{tab:sota_accuracy}
\end{table*}

Zero-shot inference yields only modest improvements over random selection, although performance slightly increases when prompts include explicit semantic cues such as ``green,” ``blue,” or ``percentage.” In contrast, fine-tuning produces substantial performance gains of approximately 90$\%$, highlighting the strong potential of large language models for accurate and scalable quantitative estimation. In particular, metrics such as Green Area Percentage consistently achieve the highest accuracy, likely due to their clear visual semantics and well-aligned linguistic representations, which facilitate both perception and numerical decoding. The presence of structured terms such as ``percentage" further strengthens model performance by providing explicit quantitative targets. While more complex metrics, such as Blue–Green Ratio, remain relatively challenging due to their reliance on higher-level relational reasoning, the overall improvements after fine-tuning indicate that these limitations could be alleviated. Collectively, these results demonstrate that, with appropriate adaptation, large language models can effectively capture and scale quantitative understanding of the built environment.

Another important observation is that current state-of-the-art remote sensing models do not consistently outperform general-purpose models on these tasks. This gap appears to stem from limitations in existing tuning strategies for domain knowledge injection. Despite being tailored for remote sensing imagery, these models still lack sufficient domain-specific understanding for quantitative reasoning over built environment metrics, particularly when interpreting spatial patterns from aerial views. This finding highlights the need for more effective mechanisms to incorporate structured domain knowledge, especially for quantitative analysis. Improving such integration is essential for advancing model performance in built environment understanding and enabling more reliable large-scale applications.

\section{Discussion}
We demonstrate that large language models can effectively estimate built environment characteristics in a quantitative manner, offering promising support for energy planning and management at scale. Most metrics show substantial improvement after domain-specific fine-tuning, highlighting the benefits of incorporating domain knowledge. While these results are encouraging and open new directions for further optimization, several limitations remain. Our analysis across cities reveals noticeable spatial variability in performance; although some larger cities exhibit higher accuracy, no consistent relationship is observed between city size and model performance. We attribute this variation primarily to inconsistencies in the quality of land use data from OpenStreetMap. In particular, cities with lower accuracy often correspond to regions with incomplete or less reliable OSM annotations, which introduce noise into derived metrics and weaken the alignment between remote sensing imagery and urban indicators. These findings suggest that data quality is critical in model performance. Therefore, improving the completeness and reliability of OSM data is essential for advancing vision–language models in built environment reasoning. In addition, planning and analytical applications based on such data should explicitly account for these limitations and incorporate strategies for data validation and quality enhancement.



\section{Conclusions}


This work develops a scalable benchmark to evaluate state-of-the-art vision language models on quantitative visual question answering for remote sensing imagery, with the goal of demonstrating the potential of large language models for quantitative understanding of the built environment and energy efficient planning. The benchmark includes a large collection of images and question–answer pairs, emphasizing challenging and practical tasks for urban analysis. Our evaluation shows that general-purpose models struggle with remote sensing recognition and numerical estimation under zero-shot settings, while domain-specific models provide only modest improvements. With targeted fine-tuning using limited data, performance improves substantially, highlighting the effectiveness of data-efficient adaptation. Overall, the results demonstrate that, with appropriate adaptation, large language models can support scalable and reliable quantitative analysis of the built environment.

\section{Acknowledgements}
This research was supported by the University of Florida President’s Strategic Initiative for AI-Enabled Smart Cities through the College of Design, Construction and Planning (DCP).

\printbibliography

@inproceedings{kuckreja_geochat_2024,
	title = {Geochat: Grounded large vision-language model for remote sensing},
	pages = {27831--27840},
	booktitle = {Proceedings of the {IEEE}/{CVF} Conference on Computer Vision and Pattern Recognition},
	author = {Kuckreja, Kartik and Danish, Muhammad Sohail and Naseer, Muzammal and Das, Abhijit and Khan, Salman and Khan, Fahad Shahbaz},
	date = {2024},
}

@inproceedings{pang_vhm_2025,
	title = {Vhm: Versatile and honest vision language model for remote sensing image analysis},
	volume = {39},
	pages = {6381--6388},
	booktitle = {Proceedings of the {AAAI} Conference on Artificial Intelligence},
	author = {Pang, Chao and Weng, Xingxing and Wu, Jiang and Li, Jiayu and Liu, Yi and Sun, Jiaxing and Li, Weijia and Wang, Shuai and Feng, Litong and Xia, Gui-Song and {others}},
	date = {2025},
	note = {Issue: 6},
}

@article{yao_falcon_2025,
	title = {Falcon: A remote sensing vision-language foundation model},
	journaltitle = {{arXiv} preprint {arXiv}:2503.11070},
	author = {Yao, Kelu and Xu, Nuo and Yang, Rong and Xu, Yingying and Gao, Zhuoyan and Kitrungrotsakul, Titinunt and Ren, Yi and Zhang, Pu and Wang, Jin and Wei, Ning and {others}},
	date = {2025},
}

@article{wang_geollava-8k_2025,
	title = {{GeoLLaVA}-8K: Scaling Remote-Sensing Multimodal Large Language Models to 8K Resolution},
	journaltitle = {{arXiv} preprint {arXiv}:2505.21375},
	author = {Wang, Fengxiang and Chen, Mingshuo and Li, Yueying and Wang, Di and Wang, Haotian and Guo, Zonghao and Wang, Zefan and Shan, Boqi and Lan, Long and Wang, Yulin and {others}},
	date = {2025},
}

@inproceedings{liu_improved_2024,
	title = {Improved Baselines with Visual Instruction Tuning},
	pages = {26296--26306},
	booktitle = {Proceedings of the {IEEE}/{CVF} Conference on Computer Vision and Pattern Recognition ({CVPR})},
	author = {Liu, Haotian and Li, Chunyuan and Li, Yuheng and Lee, Yong Jae},
	date = {2024-06},
}

@inproceedings{chen_internvl_2024,
	title = {Internvl: Scaling up vision foundation models and aligning for generic visual-linguistic tasks},
	pages = {24185--24198},
	booktitle = {Proceedings of the {IEEE}/{CVF} conference on computer vision and pattern recognition},
	author = {Chen, Zhe and Wu, Jiannan and Wang, Wenhai and Su, Weijie and Chen, Guo and Xing, Sen and Zhong, Muyan and Zhang, Qinglong and Zhu, Xizhou and Lu, Lewei and {others}},
	date = {2024},
}

@misc{OpenStreetMap,
   author = {{OpenStreetMap contributors}},
   title = {{Planet dump retrieved from https://planet.osm.org }},
   howpublished = "\url{ https://www.openstreetmap.org }",
   year = {2017},
 }

@incollection{rzeszewski2023mapbox,
  title={Mapbox},
  author={Rzeszewski, Micha{\l}},
  booktitle={Evaluating participatory mapping software},
  pages={21--40},
  year={2023},
  publisher={Springer}
}

@book{oliveira2020urban,
  title={Urban morphology},
  author={Oliveira, V{\'\i}tor},
  year={2020},
  publisher={Springer}
}

@article{yao2025falcon,
  title={Falcon: A remote sensing vision-language foundation model},
  author={Yao, Kelu and Xu, Nuo and Yang, Rong and Xu, Yingying and Gao, Zhuoyan and Kitrungrotsakul, Titinunt and Ren, Yi and Zhang, Pu and Wang, Jin and Wei, Ning and others},
  journal={arXiv e-prints},
  pages={arXiv--2503},
  year={2025}
}

@article{touvron2023llama,
  title={Llama: Open and efficient foundation language models},
  author={Touvron, Hugo and Lavril, Thibaut and Izacard, Gautier and Martinet, Xavier and Lachaux, Marie-Anne and Lacroix, Timoth{\'e}e and Rozi{\`e}re, Baptiste and Goyal, Naman and Hambro, Eric and Azhar, Faisal and others},
  journal={arXiv preprint arXiv:2302.13971},
  year={2023}
}

@article{bai2023qwen,
  title={Qwen technical report},
  author={Bai, Jinze and Bai, Shuai and Chu, Yunfei and Cui, Zeyu and Dang, Kai and Deng, Xiaodong and Fan, Yang and Ge, Wenbin and Han, Yu and Huang, Fei and others},
  journal={arXiv preprint arXiv:2309.16609},
  year={2023}
}

@inproceedings{chen2024internvl,
  title={Internvl: Scaling up vision foundation models and aligning for generic visual-linguistic tasks},
  author={Chen, Zhe and Wu, Jiannan and Wang, Wenhai and Su, Weijie and Chen, Guo and Xing, Sen and Zhong, Muyan and Zhang, Qinglong and Zhu, Xizhou and Lu, Lewei and others},
  booktitle={Proceedings of the IEEE/CVF conference on computer vision and pattern recognition},
  pages={24185--24198},
  year={2024}
}

@article{bai2025qwen3,
  title={Qwen3-vl technical report},
  author={Bai, Shuai and Cai, Yuxuan and Chen, Ruizhe and Chen, Keqin and Chen, Xionghui and Cheng, Zesen and Deng, Lianghao and Ding, Wei and Gao, Chang and Ge, Chunjiang and others},
  journal={arXiv preprint arXiv:2511.21631},
  year={2025}
}

@article{santhanam2022quantification,
  title={Quantification of green-blue ratios, impervious surface area and pace of urbanisation for sustainable management of urban lake--land zones in India-a case study from Bengaluru city},
  author={Santhanam, Harini and Majumdar, Rudrodip},
  journal={Journal of Urban Management},
  volume={11},
  number={3},
  pages={310--320},
  year={2022},
  publisher={Elsevier}
}

@article{li2025mini,
  title={Mini-gemini: Mining the potential of multi-modality vision language models},
  author={Li, Yanwei and Zhang, Yuechen and Wang, Chengyao and Zhong, Zhisheng and Chen, Yixin and Chu, Ruihang and Liu, Shaoteng and Jia, Jiaya},
  journal={IEEE Transactions on Pattern Analysis and Machine Intelligence},
  year={2025},
  publisher={IEEE}
}

@article{fathi2020machine,
  title={Machine learning applications in urban building energy performance forecasting: A systematic review},
  author={Fathi, Soheil and Srinivasan, Ravi and Fenner, Andriel and Fathi, Sahand},
  journal={Renewable and Sustainable Energy Reviews},
  volume={133},
  pages={110287},
  year={2020},
  publisher={Elsevier}
}

@article{zhang2026multi,
  title={Multi-scenario assessment of living-wall effects on building air purification, noise attenuation, and thermal insulation using quantum machine learning},
  author={Zhang, He and Srinivasan, Ravi and Chen, Chen and Yang, Xu and Ganesan, Vikram and Chen, Muyan},
  journal={Building and Environment},
  pages={114492},
  year={2026},
  publisher={Elsevier}
}

@article{im2022impact,
  title={The impact of climate change on a university campus’ energy use: Use of machine learning and building characteristics},
  author={Im, Haekyung and Srinivasan, Ravi S and Maxwell, Daniel and Steiner, Ruth L and Karmakar, Sayar},
  journal={Buildings},
  volume={12},
  number={2},
  pages={108},
  year={2022},
  publisher={MDPI}
}

@article{wang2018random,
  title={Random Forest based hourly building energy prediction},
  author={Wang, Zeyu and Wang, Yueren and Zeng, Ruochen and Srinivasan, Ravi S and Ahrentzen, Sherry},
  journal={Energy and Buildings},
  volume={171},
  pages={11--25},
  year={2018},
  publisher={Elsevier}
}

@article{li2024vision,
  title={Vision-language models in remote sensing: Current progress and future trends},
  author={Li, Xiang and Wen, Congcong and Hu, Yuan and Yuan, Zhenghang and Zhu, Xiao Xiang},
  journal={IEEE Geoscience and Remote Sensing Magazine},
  volume={12},
  number={2},
  pages={32--66},
  year={2024},
  publisher={IEEE}
}

@article{wang2025generative,
  title={Generative AI for urban planning: Synthesizing satellite imagery via diffusion models},
  author={Wang, Qingyi and Liang, Yuebing and Zheng, Yunhan and Xu, Kaiyuan and Zhao, Jinhua and Wang, Shenhao},
  journal={Computers, Environment and Urban Systems},
  volume={122},
  pages={102339},
  year={2025},
  publisher={Elsevier}
}

\end{document}